\documentclass{llncs}
\usepackage{graphicx}
\usepackage{booktabs}
\usepackage{amsmath}
\usepackage{amsfonts}
\usepackage{amssymb}
\usepackage[numbers,sectionbib]{natbib}
\newcommand{\mycite}[1]{\citeauthor{#1}~(\citeyear{#1})}
\newcommand{\myciteb}[1]{(\citeauthor{#1},~\citeyear{#1})}

\DeclareMathOperator{\rank}{rank}

\DeclareMathOperator*{\argmin}{arg\,min}

\begin{document}
\title{ Low-Rank Approximation of Matrices for PMI-based Word Embeddings}
%
%
\author{Alena Sorokina \and
Aidana Karipbayeva \and
Zhenisbek Assylbekov}
\authorrunning{Sorokina et al.}
%
\institute{Department of Mathematics, Nazarbayev University 
\\\email{\{alena.sorokina, aidana.karipbayeva, zhassylbekov\}@nu.edu.kz}}
\maketitle              
\begin{abstract}
We perform an empirical evaluation of several methods of low-rank approximation in the problem of obtaining PMI-based word embeddings. All word vectors were trained on parts of a large corpus extracted from English Wikipedia (enwik9) which was divided into two equal-sized datasets, from which PMI matrices were obtained. A  repeated measures design was used in assigning a method of low-rank approximation (SVD, NMF, QR) and a dimensionality of the vectors (250, 500) to each of the PMI matrix replicates. Our experiments show that word vectors obtained from the truncated SVD achieve the best performance on two downstream tasks, similarity and analogy, compare to the other two low-rank approximation methods.

\keywords{natural language processing \and pointwise mutual information \and matrix factorization \and low-rank approximation \and word vectors}
\end{abstract}
\section{Introduction}
Today word embeddings play an important role in many natural language processing tasks, from predictive language models and machine translation to image annotation and question answering, where they are usually `plugged in' to a larger model. An understanding of their properties is of interest as it may allow the development of better performing embeddings and improved interpretability of models using them. One of the widely-used word embedding models is the Skip-gram with negative sampling (SGNS) of \mycite{mikolov2013distributed}. \mycite{levy2014neural} showed that the SGNS is implicitly factorizing a pointwise mutual information (PMI) matrix shifted by a global constant. They also showed that a low-rank approximation of the PMI matrix by truncated singular-value decomposition (SVD) can produce word vectors that are comparable to those of SGNS. However, truncated SVD is not the only way of finding a low-rank approximation of a matrix. It is optimal in the sense that it minimizes the approximation error in the Frobenius and the 2- norms, but this does not mean that it produces optimal word embeddings, which are usually evaluated in downstream NLP tasks. The question is: Is there any other method of low-rank matrix approximation that produces word embeddings better than the truncated SVD factorization? Our experiments show that the truncated SVD is actually a strong baseline which we failed to beat by another two widely-used low-rank approximation methods.

\section{ Low-Rank Approximations of the PMI-matrix}
The simplest version of a PMI matrix is a symmetric matrix with each row and column indexed by words\footnote{Assume that words have already been converted into integer indices.}, and with elements defined as 
\begin{equation}
\mathrm{PMI}(i,j)=\log\frac{p(i,j)}{p(i)p(j)},\label{pmi}
\end{equation}
where $p(i, j)$ is the probability that the words $i$, $j$ appear within a window of a certain size in a large corpus, and $p(i)$ is the unigram probability for the word $i$. For computational purposes, \mycite{levy2014neural} suggest using a positive PMI (PPMI), defined as 
\begin{equation}
\mathrm{PPMI}(i,j)=\max(\mathrm{PMI}(i,j),0).\label{ppmi}
\end{equation}
They also show empirically that the low-rank SVD of the PPMI produces word vectors which are comparable in quality to those of the SGNS. 

The low-rank matrix approximation is approximating a matrix by one whose rank is less than that of the original matrix. The goal of this is to obtain a more compact representation of the data with a limited loss of information. In what follows we give a brief overview of the low-rank approximation methods used in our work. Since both PMI \eqref{pmi} and PPMI \eqref{ppmi} are square matrices, we will consider approximation of square matrices. For a thorough and up-to-date review of low-rank approximation methods see the paper by \mycite{kishore2017literature}.

\subsubsection{Singular Value Decomposition (SVD)} factorizes $\mathbf{A}\in\mathbb{R}^{n\times n}$, into the matrices $\mathbf{U}\in\mathbb{R}^{n\times n}$, $\mathbf{S}\in\mathbb{R}^{n\times n}$ and $\mathbf{V}^\top\in\mathbb{R}^{n\times n}$: 
$$\mathbf{A}=\mathbf{USV}^\top,$$ 
where $\mathbf{U}$ and $\mathbf{V}$ are orthogonal matrices, and $\mathbf{S}$ is a rectangular diagonal matrix whose entries are in descending order, $\sigma_1\ge\sigma_2\ge\cdots\ge\sigma_n\ge0$, along the main diagonal, and are known as the {singular values} of $\mathbf{A}$. The rank $d$ approximation (also called \textit{truncated} or \textit{partial SVD}) of $\mathbf{A}$, $\mathbf{A}_d$ where $d<\rank\mathbf{A}$, is given by zeroing out the $n-d$ trailing singular values of $\mathbf{A}$, that is\footnote{$\mathbf{A}_{a:b,c:d}$ is a submatrix located at the intersection of rows $a, a+1, \ldots, b$ and columns
$c, c + 1, \ldots, d$ of a matrix $\mathbf{A}$.}
$$\mathbf{A}_d=\mathbf{U}_{1:n,1:d}\mathbf{S}_{1:d,1:d}\mathbf{V}^\top_{1:d,1:n}.$$ 
By the Eckart-Young theorem \myciteb{eckart1936approximation}, $A_d$ is the closest rank-$d$ matrix to $A$ in Frobenius norm, i.e. $\|\mathbf{A}-\mathbf{A}_d\|_F\le\|\mathbf{A}-\mathbf{B}\|_F$, $\forall\mathbf{B}\in\mathbb{R}^{n\times n}:\,\rank(\mathbf{B})=d$.
\mycite{levy2014neural} suggest factorizing the PPMI matrix with truncated SVD, and then taking the rows of $\mathbf{U}_{1:n,1:d}\mathbf{S}_{1:d,1:d}^{1/2}$ as word vectors, and we follow their approach.

\subsubsection{QR decomposition} with column pivoting of $\mathbf{A}\in\mathbb{R}^{n\times n}$ has the form $\mathbf{A P}=\mathbf{Q  R}$, where  $\mathbf{Q}\in\mathbb{R}^{n\times n}$ is orthogonal, $\mathbf{R}\in\mathbb{R}^{n\times n}$ is upper triangular and  $\mathbf{P}\in\mathbb{R}^{n\times n}$ is a permutation matrix. The rank $d$ approximation to $\mathbf{A}$ is then $$\mathbf{A}_d=\mathbf{Q}_{1:n,1:d}[\mathbf{RP}^\top]_{1:d,1:n}$$ 
which is called \textit{truncated QR decomposition} of $\mathbf{A}$. After factorizing the PPMI matrix with this method we suggest taking the rows of $\mathbf{Q}_{1:n,1:d}$ as word vectors. 

However, we suspect that a valuable information could be left in the $\mathbf{R}$ matrix. A promising alternative to SVD is a Rank Reveling QR decomposition (RRQR). Assume the QR factorization of the matrix $\mathbf{A}$:
\[
   \mathbf{AP}= \mathbf{Q}\left[ {\begin{array}{cc}
  \mathbf{R}_{11} & \mathbf{R}_{12} \\
   0 & \mathbf{R}_{22} \\
  \end{array} } \right]
\]
where $\mathbf{R}_{11}\in\mathbb{R}^{d\times d}$, $\mathbf{R}_{12}\in\mathbb{R}^{d\times (n-d)}$, $\mathbf{R}_{22}\in\mathbb{R}^{(n-d)\times (n-d)}$. For RRQR factorization, the following
condition should be satisfied: 
\begin{align*}
\sigma_{\min}(\mathbf{R}_{11}) &= \mathrm{\Theta}(\sigma_{d}(\mathbf{A}))\\
\sigma_{\max}(\mathbf{R}_{22}) &= \mathrm{\Theta}(\sigma_{d+1}(\mathbf{A}))
\end{align*}
which suggests that the most significant entries are in $\mathbf{R}_{11}$, and the least important are in $\mathbf{R}_{22}$.  Thus, we also suggest taking the columns of  $[\mathbf{RP}^\top]_{1:d,1:n}$ as word vectors.

\subsubsection{Non negative matrix factorization (NMF).} Given a non negative matrix $\mathbf{A}\in\mathbb{R}^{n\times n}$ and a positive integer $d<n$, NMF finds non negative matrices $\mathbf{W}\in\mathbb{R}^{n\times d}$ and $\mathbf{H}\in\mathbb{R}^{d\times n}$ which  minimize (locally) the functional $f(\mathbf{W},\mathbf{H})=\|\mathbf{A}-\mathbf{WH}\|^2_F$. The rank $d$ approximation of $\mathbf{A}$ is simply
$$
\mathbf{A}_d=\mathbf{WH}.
$$
When factorizing the PPMI matrix with NMF, we suggest taking the rows of $\mathbf{W}$ as word vectors.

\section{Experimental Setup}
\subsection{Corpus} 
All word  vectors were trained on the \texttt{enwik9} dataset\footnote{\url{http://mattmahoney.net/dc/textdata.html}} which was divided into two equal-sized splits. The PMI matrices on these splits were obtained using the \texttt{hypewords} tool of \mycite{levy2015improving}. All corpora were pre-processed by removing non-textual elements, sentence splitting, and tokenization. PMI matrices were derived using a window of two tokens to each side of the focus word, ignoring words that appeared less than 300 times in the corpus, resulting in vocabulary sizes of roughly 13000 for both words and contexts. A  repeated measures design was used for assigning the method of factorization (SVD, QR, NMF) and dimensionality of the vectors (250, 500) to each PMI matrix replicate. We used two replicates per each level combination.

\subsection{Training} 
Low-rank approximations were performed using the following open-source implementations: 
\begin{itemize}
    \item Sparse SVD from SciPy \myciteb{jones2014scipy},
    \item Sparse RRQR from SuiteSparse \myciteb{davis2011university}, and
    \item NMF from scikit-learn \myciteb{pedregosa2011scikit}.
\end{itemize}
For NMF we used the nonnegative double SVD initialization. We trained 250 and 500 dimensional word vectors with each method. 

\subsection{Evaluation} 
We evaluate word vectors on two tasks: similarity and analogy. A similarity is tested using the WordSim353 dataset of \mycite{finkelstein2002placing}, containing word pairs with human-assigned similarity scores. Each word pair is ranked by cosine similarity and the evaluation is the Spearman correlation between those rankings and human ratings. Analogies are tested using Mixed dataset of 19544 questions such as ``$a$ is to $b$ as $c$ is to $d$'', where $d$ is hidden and must be guessed from the entire vocabulary. We filter questions with out of vocabulary words, as standard. Accuracy is computed by comparing $\argmin_{d}\|\mathbf{b} - \mathbf{a} + \mathbf{c} - \mathbf{d}\|$ to the labelled answer.

\section{Results} The results of evaluation are provided in Table~\ref{tab1}, which we analyze using the two-factor ANOVA with factors being (1) low-rank approximation method and (2) dimensinality of word vectors, and response being the performance in similarity or analogy task. We analyze the tasks separately.
\setlength{\tabcolsep}{8pt}
\begin{table}
\begin{center}
\caption{Results}\label{tab1}
\begin{tabular}{c c c c c}
\toprule
Method of low- & Dimensionality & Replicate & Similarity & Analogy \\
rank approximation & of vectors & \# & task & task\\
\midrule
SVD &  250 & 1 & 0.7010 & 0.3778\\
SVD &  250 & 2 & 0.6969 & 0.3817\\
SVD &  500 & 1 & 0.6989 & 0.3568\\
SVD &  500 & 2 & 0.6914 & 0.3458\\
NMF &  250 & 1 & 0.5265 & 0.0660\\
NMF &  250 & 2 & 0.4780 & 0.0563\\
NMF &  500 & 1 & 0.4499 & 0.0486\\
NMF &  500 & 2 & 0.3769 & 0.0487\\
QR (R) &  250 & 1 & 0.4077 & 0.1644\\
QR (R) &  250 & 2 & 0.3822 & 0.1533\\
QR (R) &  500 & 1 & 0.4717 & 0.2284\\
QR (R) &  500 & 2 & 0.3719 & 0.1925\\
QR (Q)&  250 & 1 & 0.2870 & 0.0034\\
QR (Q)&  250 & 2 & 0.2009 & 0.0059\\
QR (Q)&  500 & 1 & 0.3573 & 0.0165\\
QR (Q)&  500 & 2 & 0.3048 & 0.0186\\
\bottomrule
\end{tabular}
\end{center}
\end{table}

\begin{figure}
\centering
\includegraphics[width=.49\textwidth]{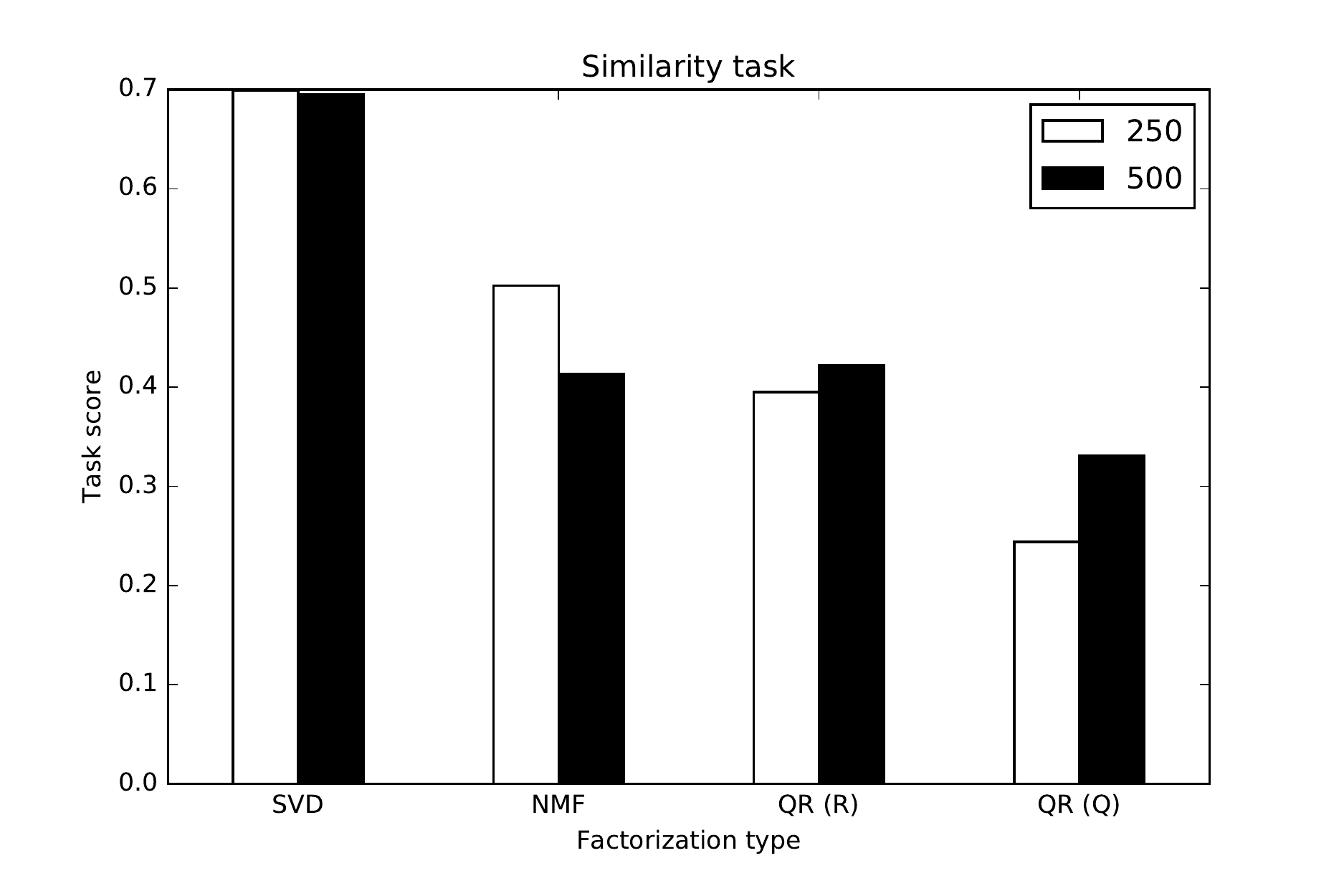}
\includegraphics[width=.49\textwidth]{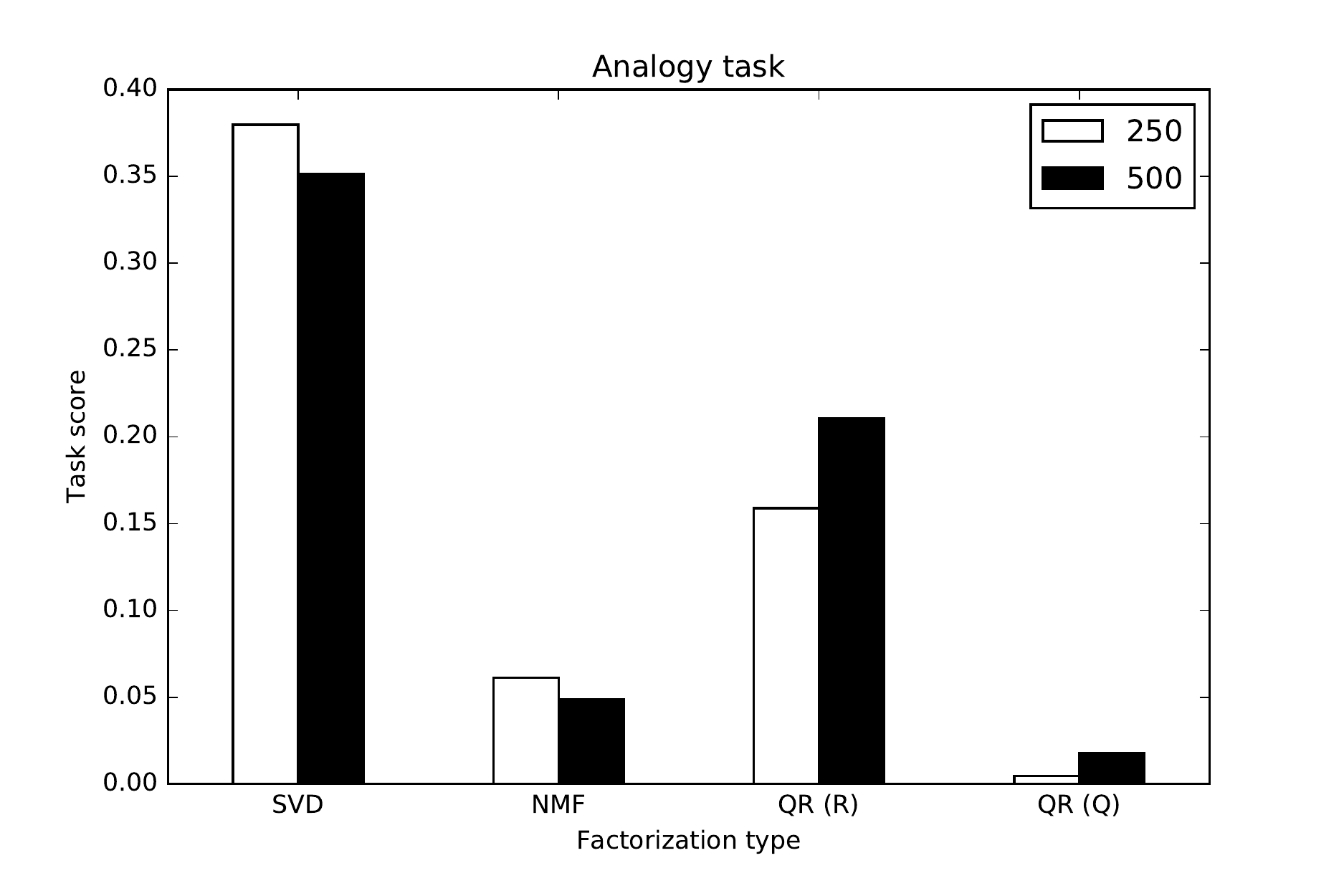}
\caption{Test scores for different factorization methods on Similarity and Analogy tasks.}

\label{residuals1}
\end{figure}

\subsection{Similarity task} 

\begin{figure}
\centering
\includegraphics[height=.45\textheight]{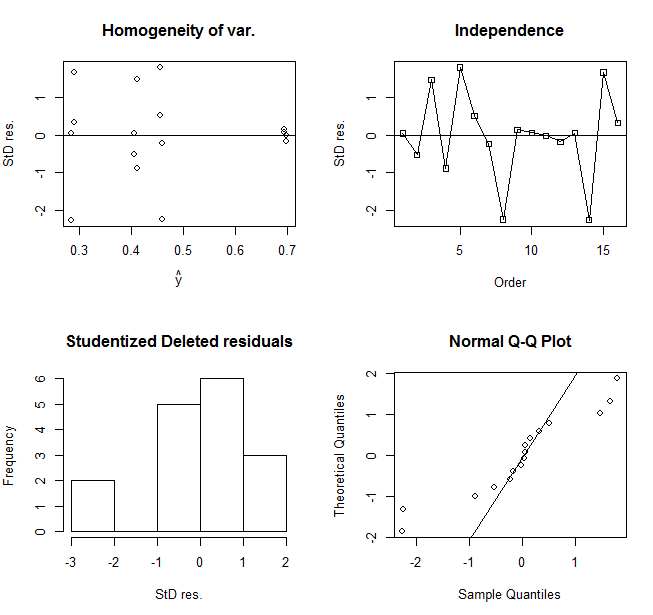}
\caption{ANOVA residuals for the results on Similarity task.}

\label{residuals1}
\end{figure}

The standard residual analysis is used to check whether the ANOVA assumptions are satisfied. From  Figure~\ref{residuals1} we see that the residuals have constant variability around zero, are independent and normally distributed. The normality is confirmed using Shapiro-Wilk test, $p\text{-value} = 0.7923$. 
\setlength{\tabcolsep}{4pt}
\begin{table}
\begin{center}
\caption{ANOVA table for the similarity task results}\label{tab2}
\begin{tabular}{l c c  c c c}
\toprule
Source & DF & Sum of Squares & Mean Square & F Value & Pr ${>}$ F\\
\midrule
Model &  7 & 0.37055017 & 0.05293574 & 29.68 & ${<}$ .0001\\
Error &  8 & 0.01426728 & 0.00178341 & & \\
Corrected total &  15 & 0.38481745 & & &\\
\midrule
 & R-Square & Coeff Var & Root MSE & Score Mean \\
 & 0.962925 &  9.126819 & 0.042230 & 0.462707 \\
\bottomrule
\end{tabular}
\end{center}
\end{table}
\begin{table}
\begin{center}
\caption{Main and Interaction Effects in the Similarity task}\label{tab3}
\begin{tabular}{l c c c c c}
\toprule
Source & DF & Sum of Squares & Mean Square & F Value & Pr ${>}$ F\\
\midrule
Factorization &  3 & 0.35433596 & 0.11811199 & 66.23 & ${<}$ .0001\\
Dimension &  1 & 0.00011159 & 0.00011159 & 0.06 & 0.8088\\
Interaction &  3 & 0.01610263 & 0.00536754 & 3.01 & 0.0945\\
\bottomrule
\end{tabular}
\end{center}
\end{table}

\begin{figure}[htbp]
\centering
\includegraphics[height=.45\textheight]{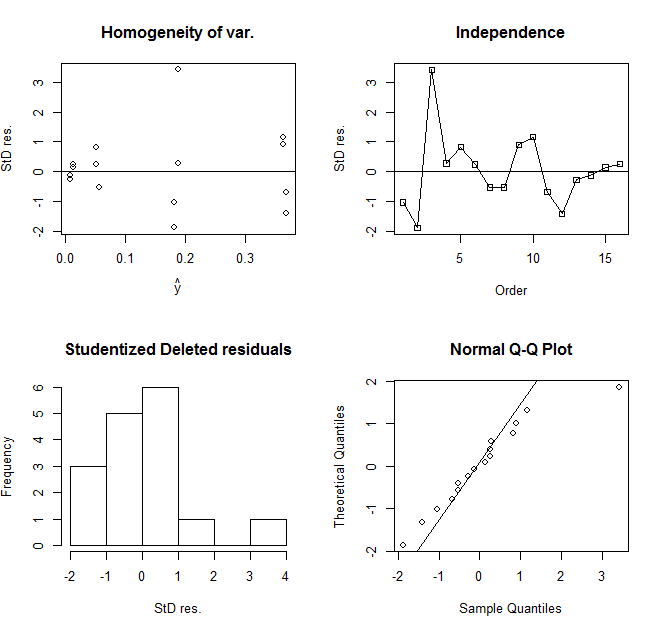}
\caption{ANOVA Residuals for the Analogy task results}
\label{fig2}
\end{figure}
\begin{table}[htbp]
\begin{center}
\caption{ANOVA Table for the Analogy task results}
\label{tab4}
\begin{tabular}{l c c c c c}
\toprule
Source & DF & Sum of Squares & Mean Square & F Value & Pr ${>}$ F\\
\midrule
Model &  7 & 0.30745304 & 0.04392186 & 424.61 & ${<}$ .0001\\
Error &  8 & 0.00082753 & 0.00010344 & & \\
Corrected total &  15 & 0.30828057 & & &\\
\midrule
 & R-Square & Coeff Var & Root MSE & Score Mean \\
 & 0.997316 & 6.602449 & 0.010171 & 0.154043 \\
\bottomrule
\end{tabular}
\end{center}
\end{table}
\begin{table}[htbp]
\begin{center}
\caption{Main and Interaction Effects in the Analogy task}\label{tab5}
\begin{tabular}{l c c c c c}
\toprule
Source & DF & Sum of Squares & Mean Square & F Value & Pr ${>}$ F\\
\midrule
Factorization &  3 & 0.30365768 & 0.10121923 & 978.52 & ${<}$ .0001\\
Dimension &  1 & 0.00013820 & 0.00013820 & 1.34 & 0.2811\\
Interaction &  3 & 0.00365715 & 0.00121905 & 11.78 & 0.0026\\
\bottomrule
\end{tabular}
\end{center}
\end{table}

The SAS package was used to obtain ANOVA table (Table~\ref{tab2}), which shows the effects of the factors on the similarity score.  F-test for equality of the factor level means was conducted, $F = 29.68$ and $\text{p-value}<0.0001$. Hence, it can be concluded that at least one factor level mean is different from the others. ${R}^2 = 0.962925$ shows that more than 96\% of variation in the similarity score is explained by the factors considered.

Proceeding with analysis of main and interaction effects, one can conduct F-test for each of the factors and the interaction between them. From  Table~\ref{tab3}, we see that the method of low-rank approximation  affects the performance of words vectors in the similarity task, $F = 66.23$, p-value $< 0.0001$. The dimensionality of word vectors has no effect on the performance in the similarity task, $F = 0.06$ with $\text{p-value} > 0.8$. Also, there is no interaction between the method of factorization and the dimensionality of word vectors, $F = 3.01$ with p-value $0.0945$. Thus, SVD significantly outperforms the other factorization methods.

\subsection{Analogy task} 
Again, we first need to check whether the ANOVA assumptions are satisfied. From  Figure~\ref{fig2} we see that the residuals have constant variability around zero, are independent and normally distributed. The normality is confirmed using Shapiro-Wilk test, $\text{p-value} = 0.112$.
The ANOVA Table (Table ~\ref{tab4}) shows that at least one level mean is different from the others. ${R}^2$ is 0.997316, thus, 99\% of variation in the analogy score is explained by the considered factors.

We proceed to the  analysis  of  main  and  interaction effects. The method of low-rank approximation  affects the performance of word vectors in the analogy task, $F = 978.52$ with $\text{p-value} < 0.0001$. The dimensionality of word vectors has no effect on the performance in the analogy task, $F = 1.34$ with $\text{p-value} > 0.2$. Unlike the similarity task, there is an interaction effect between the two factors, $F = 11.78$ with $\text{p-value} = 0.0026$.

\section{Discussion}

\noindent\textbf{Why dimensionality is critical in similarity task for NMF?}
We obtained the highest results in the similarity task using the SVD-based low-rank approximation, for which the dimensionality of word vectors did not influence the performance much. On the contrary, the performance in similarity task using the NMF method of factorization is significantly affected by the dimension of the word vector: 250-dimensional word vectors give significantly better results than 500-dimensional ones. This can be explained by the specific characteristics of the NMF method of factorization. When we look at the word vectors produced by NMF, we can see that they contain many zeros. Hence, an increase in the dimensionality makes them even sparser. Similarity task is based on finding the cosine of the angle between two word vectors. Therefore, when the vectors become sparser, the result of element-wise multiplication, which is necessary for obtaining cosine, becomes smaller. Thus, there is a much higher possibility that the cosine similarity score between two vectors, containing many zeros, will give a number closer to zero than to 1. This, as a result, leads to the worse performance in the similarity task. Our suggestion is to decrease the dimensionality of the NMF method to 100. We expect that this may give better results.
\newline

\noindent\textbf{Why NMF performs poorly in the analogy task?}
We provide a theoretical analysis of the poor performance of the NMF in the analogy task. We model word vectors produced by the NMF as independent and identically distributed random vectors from an isotropic multivariate Gaussian distribution $\mathcal{N}(\mathbf{4.5},\mathbf{I})$\footnote{The isotropy is motivated by the work of \mycite{arora2016latent}; $\mathbf{4.5}$ is a vector with all elements equal to $4.5$.}, since for a 500-dimensional $\mathbf{v}\sim\mathcal{N}(\mathbf{4.5},\mathbf{I})$ there is a big chance that it is nonnegative:
$$
\Pr(\mathbf{v}\in[0,+\infty)^{500})=[\Pr(4.5+Z>0)]^{500}\approx0.9983,
$$
where $Z\sim\mathcal{N}(0,1)$ is a standard normal random variable. For a triplet of word vectors $\mathbf{a}$, $\mathbf{b}$ and $\mathbf{c}$ we have $\mathbf{b}-\mathbf{a}+\mathbf{c}\sim\mathcal{N}(\mathbf{4.5},3\mathbf{I})$, and therefore
\begin{align*}
&\Pr(\mathbf{b}-\mathbf{a}+\mathbf{c}\in[0,+\infty)^d)=[\Pr(3+\sqrt{3}Z\ge0)]^d\\
&=[\Pr(Z\ge-4.5/\sqrt{3})]^d<[0.9953]^d.
\end{align*}
When $d=500$, this probability is $\approx0.1$, i.e. there is a small chance that $\mathbf{b}-\mathbf{a}+\mathbf{c}$ is non negative, and thus we will likely not find a non-negative $\mathbf{d}$ when we minimize $\|\mathbf{b}-\mathbf{a}+\mathbf{c}-\mathbf{d}\|$. This is confirmed empirically: for \textit{all} word triplets $(a,b,c)$ from the analogy task, the vector $\mathbf{b}-\mathbf{a}+\mathbf{c}$ has at least one negative component.
\newline

\noindent\textbf{Why using $\mathbf{R}$ is better than using $\mathbf{Q}$ in the QR decomposition?}
The $\mathbf{Q}$ matrix from QR factorization gives the worst results in the similarity task, and it does not depend on the dimensionality of the vector. The reason is that the necessary information is left in the $\mathbf{R}$ matrix.   Truncation of $\mathbf{RP}^\top$ gives better approximation to the original matrix than the truncated $\mathbf{Q}$, because the most significant entries of $\mathbf{RP}^\top$ are in the top left quarter and remain after truncation.

\section{Conclusion}
We analyzed the performance of the word vectors obtained from a word-word PMI matrix by different low-rank approximation methods. As it was expected, the truncated SVD provides a far better solution than the NMF and the truncated QR in both similarity and analogy tasks. While the performance of the NMF is relatively good in the similarity task, it is significantly worse in the analogy task. NMF produces only non-negative sparse vectors and we showed how this deteriorates the performance in both tasks. $\mathbf{RP}^\top$ matrix from QR factorization with column pivoting gives better word embedding than $\mathbf{Q}$ matrix in both tasks.

\section*{Acknowledgement}
The work of Zhenisbek Assylbekov has been funded by the Committee of Science of the Ministry of Education and Science of the Republic of Kazakhstan, contract \# 346/018-2018/33-28, IRN AP05133700.

\bibliography{ref}

\end{document}